%% file: paper_new_8p.tex
\documentclass[letterpaper, 10pt]{article}

\usepackage[letterpaper,left=1in,top=1in,right=1in,bottom=1in,nohead]{geometry}

\usepackage{float}
\usepackage{times} 
\usepackage{multicol}
\usepackage{longtable}

\usepackage{amsmath}
\usepackage{graphicx} 
\usepackage{subfig}
\usepackage{amsfonts}
\usepackage{xcolor}
\usepackage{wrapfig}

\usepackage[noend]{algpseudocode}
\usepackage{algorithm}
\usepackage{color}
\usepackage[font=small]{caption}

\newlength{\sectiongaps}
\newcommand{\mysection}[1]{\vspace{0.8\sectiongaps}\section{{#1}}\vspace{1.2\sectiongaps}}
\newcommand{\mysubsection}[1]{\vspace{\sectiongaps}\subsection{{#1}}\vspace{\sectiongaps}}
\newcommand{\mysubsubsection}[1]{\vspace{0.4\sectiongaps}\subsubsection{{#1}}}

\allowdisplaybreaks

\newlength{\eqnpregap}
\newlength{\eqnpostgap}
\newcommand{\eqnsize}{} 
\newcommand{\algsize}{} 

\algnewcommand\algorithmicforeach{\textbf{for each}}
\algdef{S}[FOR]{ForEach}[1]{\algorithmicforeach\ #1\ \algorithmicdo}

\definecolor{MyBlue}{rgb}{0,0,1.0}
\definecolor{MyRed}{rgb}{1.0,0,0}
\definecolor{MyGrey}{rgb}{0.5,0.5,0.5}
\definecolor{MyDarkRed}{rgb}{0.5,0,0.1}
\definecolor{MyDarkBlue}{rgb}{0.1,0.1,0.5}
\definecolor{MyDarkGreen}{rgb}{0.1,0.5,0.1}

\usepackage[pdftex, colorlinks=true, urlcolor=MyDarkBlue, citecolor=MyDarkRed, linkcolor=MyDarkGreen]{hyperref}

\newcommand{\algcomment}[1]{{\color{MyGrey}{\small ~~// {#1}}}}

\newcommand{\changeJ}[1]{{#1}}
\newcommand{\changed}[1]{{#1}} 
\newcommand{\amanda}[1]{{#1}} 
\newcommand{\changedB}[1]{{#1}} 
\newcommand{\changeC}[1]{{#1}}
\newcommand{\changeD}[1]{{#1}} 
\newcommand{\changedE}[1]{{#1}} 

\DeclareMathOperator*{\argmin}{argmin}

\title{\LARGE \bf
\changeD{Pursuer Assignment and Control Strategies in \\ Multi-agent Pursuit-Evasion Under Uncertainties}
}

\author{
Leiming Zhang${}^{*}$\thanks{${}^{*}$Department of Mechanical Engineering and Mechanics, Lehigh University, 19 Memorial Drive West, Bethlehem, PA 18015, U.S.A., \texttt{[lez316,sub216]@lehigh.edu}.}, 
Amanda Prorok${}^{\dagger}$\thanks{${}^{\dagger}$Department of Computer Science and Technology, Cambridge University, 15 JJ Thomson Avenue, Cambridge CB3 0FD, UK, \texttt{asp45@cam.ac.uk}.} and 
Subhrajit Bhattacharya${}^{*}$
\vspace{-0.1in}
}

\newcommand{\evaderpos}{y}

\begin{document}

\maketitle
\thispagestyle{empty}
\pagestyle{empty}


\begin{abstract}
We consider a pursuit-evasion problem with \changed{a heterogeneous team of multiple pursuers} and multiple evaders. 
Although both the pursuers (robots) \changeD{and the evaders are aware of each others' control and assignment strategies, they do not have exact information about the other type of \changedE{agents' location or action}. Using only noisy on-board sensors the pursuers (or evaders) make probabilistic estimation of positions of the evaders (or pursuers). 
%
Each type of agent
use Markov localization to update the probability distribution of the other type.} A search-based \changeD{control} strategy is developed \changeD{for the pursuers} that intrinsically takes the probability distribution of the evaders into account.
Pursuers are assigned using an assignment algorithm that takes redundancy (i.e., an excess in the number of pursuers than the number of evaders) into account, such that the \changeJ{total or maximum} estimated time to capture the evaders is minimized.
\changeD{In this respect we assume the pursuers to have clear advantage over the evaders. However, the objective of this work is to use assignment strategies that minimize the capture time.
This assignment strategy is based on a modified Hungarian algorithm as well as a novel algorithm for determining assignment of redundant pursuers. The evaders, in order to effectively avoid the pursuers, predict the assignment based on their probabilistic knowledge of the pursuers and use a control strategy to actively move away from those pursues.}
\changedB{Our experimental evaluation shows that the redundant assignment algorithm performs better than an alternative nearest-neighbor based assignment algorithm.}
\footnote{\changeD{Some parts of this paper appeared as an extended abstract in the proceeding of the 2019 IEEE International Symposium on Multi-robot and Multi-agent Systems (MRS)~\cite{Zhang:MRS:18}.}}
\end{abstract}

\vspace{-0.05in}

\mysection{INTRODUCTION}

\changedE{
\mysubsection{Motivation}
Pursuit-evasion is an important problem in robotics with a wide range of applications including environmental monitoring and surveillance.
Very often evaders are adversarial agents whose exact locations or actions are not known and can at best be modeled stochastically. Even when the pursuers are more capable and more numerous than the evaders, capture time may be highly unpredictable in such probabilistic settings. Optimization of time-to-capture in presence of uncertainties is a challenging task, and an understanding of how best to make use of the excess resources/capabilities is key to achieving that.
This paper address the problem of assignment of pursuers to evaders and control of pursuers under such stochastic settings in order to minimize the expected time to capture.}


\mysubsection{Ralated Work}

\changedB{The} pursuit-evasion problem in a probabilistic setting requires localization of the evaders as well as development of \changedB{a} controller for the pursuer to enable it \changedB{to} capture the evader.
Markov localization is an effective approach for tracking probabilistic agents in unstructured environments since it is capable of representing probability distributions more general than normal distributions (unlike Kalman filters~\cite{Barshan:navigation:95}). Compared to Monte Carlo or particle filters~\cite{eacae7d713074eb4bd95d240ba8357c0,fox1999monte}, Markov localization is often computationally less intensive, more accurate and has stronger formal underpinnings.

Markov localization has been \changedB{widely} used for 
estimating an agent's position in known environments~\cite{Burgard:1996:EAP:1864519.1864520} and in dynamic environments~\cite{Fox:1999:MLM:3013545.3013556} using on-board sensors, as well as for localization of evaders using noisy external sensors~\cite{zhang2007probabilistic,fox1998active,fox1999markov}.
More recently, in conjunction with sensor fusion techniques, Markov localization has been used for target tracking using multiple sensors~\cite{Zhang:tracking:07,Nagaty:Localization:15}.

Detection and pursuit of an uncertain or unpredictable evader has also been studied extensively.
\cite{chung2011search} provides a taxonomy of search and pursuit \changed{problems} in mobile robotics. Different methods are compared in both graphs and polygonal environments.
\changedB{Importantly,} this survey also notes that the minimization of distance and time to capture the evaders is less studied.
\changeJ{\cite{coopmultirobotobserve} is another comprehensive review focused on cooperative multi-robot targets observation.}
\cite{hollinger2007probabilistic} describes strategies for pursuit-evasion in an indoor environment which is discretized into different cells, with each cell representing a room.  However, in our approach, the environment is discretized into finer grids that \changedB{generalize} to a wider variety of environments. 
In \cite{hespanha1999multiple} a probabilistic framework for a pursuit-evasion game with one evader and multiple pursuers is described. A game-theoretic approach is used in \cite{hespanha2000probabilistic} to describe a pursuit-evasion game in which evaders try to actively avoid the pursuers. 
\changeD{\cite{Makkapati:optimal:evader:19} describes an optimal strategy for evaders in multi-agent pursuit-evasion without uncertainties. Along similar lines, in~\cite{Oyler:obstacles:16} the authors describe a pursuit-evasion game in presence of obstacles in the environment.}
\changeJ{\cite{shkurti2018model}} describes a problem involving a robot that tries to follow a moving target using visual data. 
\changeJ{Patrolling is another approach to pursuit-evasion problems in which persistent surveillance is desired. 
Multi-robot patrolling with uncertainty have been studied extensively in \cite{agmon2009adversarial}, \cite{6224708} and \cite{ijcai2017-61}.}
%
%
\changed{More recently \changedB{in} \cite{shah2019multi},} Voronoi partitioning has been used 
to guide pursuers to maximally reduce the area of \changed{workspace} reachable by \changedB{a single} evader.
Voronoi partitioning along with area minimization has also been used for pursuer-to-evader assignments in problems involving multiple deterministic \changedB{and localized} evaders and pursuers~\cite{PiersonEtAlRAL17PursuitEvasion}.


\mysubsection{Problem Description \changeD{and Assumptions}}

We consider a multi-agent pursuit-evasion problem where, in a known environment, we have several surveillance robots (the pursuers) for monitoring a workspace for potential intruders (the evaders).\footnote{In this paper we have used the words \emph{``pursuer''} and \emph{``robot''} interchangeably.}
Each evader emits a weak and noisy signal (for example, wifi signal \changeJ{used by the evaders for communication} or infrared heat signature), that the \changeD{pursuers} can detect using noisy sensors to estimate their position and try to localize them.
\changeJ{We assume that the signals emitted by each evader are distinct and is different from any type of signal that the pursuers might be emitting. Thus the pursuers can not only distinguish between the signals from the evaders and other pursuers, but also distinguish between the signals emitted by the different evaders.}
\changeD{Likewise, each pursuer emits a distinct weak and noisy signal that the evaders can detect to localize the pursuers.
Each agent is aware of its own location in the environment and the agents of the same type (pursuers or evaders) can communicate among themselves. The environment (obstacle map) is assumed to be known to either type of agents.}

\changeD{Each evader uses a control strategy to actively avoid the robots pursuing it by choosing a velocity that takes it away from the pursuers. The pursuers use 
	a control strategy that allow them to follow the path with least expected capture time. 
The evaders and pursuers are aware of each others' strategies (this, for example, represents real-world scenario where every agent uses an open-source control algorithm), however, the exact locations and actions taken by one type of agent (evader/purser) at an instant of time is not known to the other type (pursuer/evader). Using the noisy signals and probabilistic sensor models, each type of agent maintains and updates (based on sensor measurements as well as the known control/motion strategy) a probability distribution that represents the locations of the individuals of the other type (pursuer/evader).

With the evaders being represented \changedB{by} probability distributions by the pursuers, the time-to-capture an evader by a particular pursuer is a stochastic variable. We thus consider the problems of pursuer-to-evader assignment and computation of control velocities for the pursuers \changedB{with a view of 
	minimizing} the \emph{total expected capture time} (the sum of the times taken to capture each of the evaders) or the \emph{maximum expected capture time} (the maximum out of the times taken to capture each of the evaders).
We assume that the number of pursuers \changedB{is} greater that the number of evaders and that the pursuers constitute a heterogeneous team, with each having different maximum speeds and different capture capabilities.
The speed of the pursuers are assumed to be higher than the evaders to enable capture in any environment (even obstacle-free environment).
The objective of this paper is to design strategies for the pursuers to assign themselves to the evades, and in particular, algorithms for assignment of the excess (redundant) pursuers, so as to minimize the total/maximum expected capture time.

While the evaders know the pursuers' assignment strategy, they don't know the pursuers' positions, the probability distributions that the pursuers use to represent the evaders, or the exact assignment that the evaders determine. Instead, the evaders rely on the probability distributions that they use to represent the pursuers to figure out the assignments that the pursuers are likely using.}
%
We use a Markov localization~\cite{probRob:Thrun} technique to update the probability distribution of each \changeD{agent}.

\mysubsection{Contributions}
\changeD{The main contributions of this paper are novel methods for pursuer-to-evader assignment in presence of uncertainties for total capture time minimization as well as for maximum capture time minimization. We also present a novel control algorithm for pursuers based on Theta* search~\cite{conf-aaai-NashDKF07} that \changeD{takes} the evaders' probability distribution into account, and control strategy for evaders that try to actively avoid the pursuers trying to capture it. We assume that both agents (pursuers and evaders) know each others' control strategies, and use that knowledge to predict and update the probability distributions that they use to represent the other type of agent.}

\mysubsection{Notations}

\noindent
\emph{Configuration Space Representation:} 
We consider a subset of the Euclidean \changeJ{plane}, $C \subset \mathbb{R}^{2}$, as the \changed{configuration space for the \changeD{robots (pursuers)} as well as the evaders, which we discretize into a set of cells or vertices, $V$,  where the agents can reside} (Figure~\ref{fig:eight_connect_graph}).
A vertex in $V$ will be represented with a lower-case letter $v\in V$, while its physical position (Euclidean coordinate vector) in $C$ will be represented as $\mathbf{X}(v)$.
For \changedB{simplicity}, we also use a discrete time representation.

\vspace{0.03in}\noindent
\emph{Agents:} 
\changed{The $i^{\text{th}}$ \changeD{pursuer/}robot's location is represented by $r_i \in V$, and the $j^{\text{th}}$ evader by $y_j \in V$ (we will use the same notations to refer to the respective agents themselves).
The set of the indices of all the pursuers is denoted by $\mathcal{C}_r$, and the set of the indices of all the evaders by $\mathcal{C}_y$.

\vspace{0.03in}\noindent
\emph{Heterogeneity:} Robot $r_i$ is assumed to have a maximum speed of $v_i$, and the objective being time minimization, it always maintains that highest possible speed. It also has a \emph{capture radius} (\emph{i.e.} the radius \changedB{of the disk} within which it can capture an evader) of $\rho_i$.

\vspace{0.03in}\noindent
\emph{Assignment:} The set of pursuers assigned to the $j^{\text{th}}$ evader will be represented by the set $I_j$. 
The individual assignment of $i^{\text{th}}$ pursuer to $j^{\text{th}}$ evader will be denoted by the pair $(i,j)$.
$\mathcal{F} = \{(i,j) |  i \in \mathcal{C}_{r}, j \in \mathcal{C}_{y}\}$ denotes the set of all possible such pursuer-to-evader pairings.

A (valid) assignment, $\mathcal{A} \subseteq \mathcal{F}$, is such that for every $(i,j), (i',j')\in \mathcal{A}$, we should have $i\!=\!i' \Rightarrow j \!=\! j'$ 
(\emph{i.e.}, a pursuer cannot be assigned to two different evaders).
This also implies $|\{j ~|~ (i,j) \in \mathcal{A}\}| \leq 1, ~\forall i\in\mathcal{C}_r$ (note that an assignment allows for unassigned pursuers).

The set of all possible valid assignments is denoted by $\mathfrak{A} = \{\mathcal{A} \subseteq \mathcal{F} ~|~ \forall (i,j), (i',j')\in \mathcal{A}, ~i\!=\!i' \Rightarrow j\!=\!j' \}$.
}

\mysubsection{Overview of the Paper}

In Section~\ref{sec:prob_repre}, we introduce
\changeD{the control strategies for the evaders and pursuers. In presence of uncertainties this control strategy becomes a stochastic one. We also describe how each type of agent predict and update the probability distributions representing the other type using this known control strategy.}
In Section~\ref{sec:assign}, we present \changed{an algorithm for assigning redundant pursuers to the probabilistic evaders so as to minimize the net expected time as well as maximum expected time to capture}. 
In Section~\ref{sec:res} simulation and comparison results are presented.


\changeD{ \vspace{-0.5em}
\mysection{Probabilistic Representation and Control Strategies} \label{sec:prob_repre}
\vspace{0.5em}


\mysubsection{Probabilistic Representation of the Agents}

The pursuers represent the $j^{\text{th}}$ evader by a probability distribution over $V$ denoted by $p^t_j:V \rightarrow\mathbb{R}_+$.
Likewise the evaders represent the $i^{\text{th}}$ pursuer by a probability distribution over $V$ denoted by $q^t_j:V \rightarrow\mathbb{R}_+$. 
The pursuers maintain the evader distributions, $\{p^t_j\}_{j\in \mathcal{C}_y}$, which are unknown to the evaders. While the evaders maintain the pursuer distributions, $\{q^t_i\}_{i\in \mathcal{C}_r}$, which are unknown to the pursuers.
The superscript $t$ emphasizes that the distributions are time-varying since they are updated by each type of agent (pursuer/evader) \changed{based on known} control strategy of the other type of agent (evader/pursuer)
 and models \changed{for sensors on-board the agents}.

\mysubsubsection{Motion Model}
At every time-step the known control strategy allows one type of agent to \emph{predict} the probability distribution of the other type of agent in the next time-step using a linear model:

{\eqnsize \begin{eqnarray}
\widetilde{p}_j^t (y) = \sum_{y'\in V} K_j (y,y') ~p_j^{t-1}(y') \nonumber \\
\widetilde{q}_i^t (r) = \sum_{r'\in V} L_i (r,r') ~q_i^{t-1}(r')
\end{eqnarray}}

\vspace{-1.5em} \noindent
where using the first equation the pursuers predict the $j^\text{th}$ evader's probability distribution at the next time-step using the \emph{transition probabilities} $K_j$ computed using the known control strategy of the evader.
While the second equation is used by the evaders to predict the $i^\text{th}$ pursuer's probability distribution using transition probabilities $L_i$ computed from the known control strategy of the pursuers.

\begin{wrapfigure}{r}{0.36\textwidth} \vspace{-0.2in} 
	\centering
	\includegraphics[width=0.32\textwidth, trim=30 0 30 0]{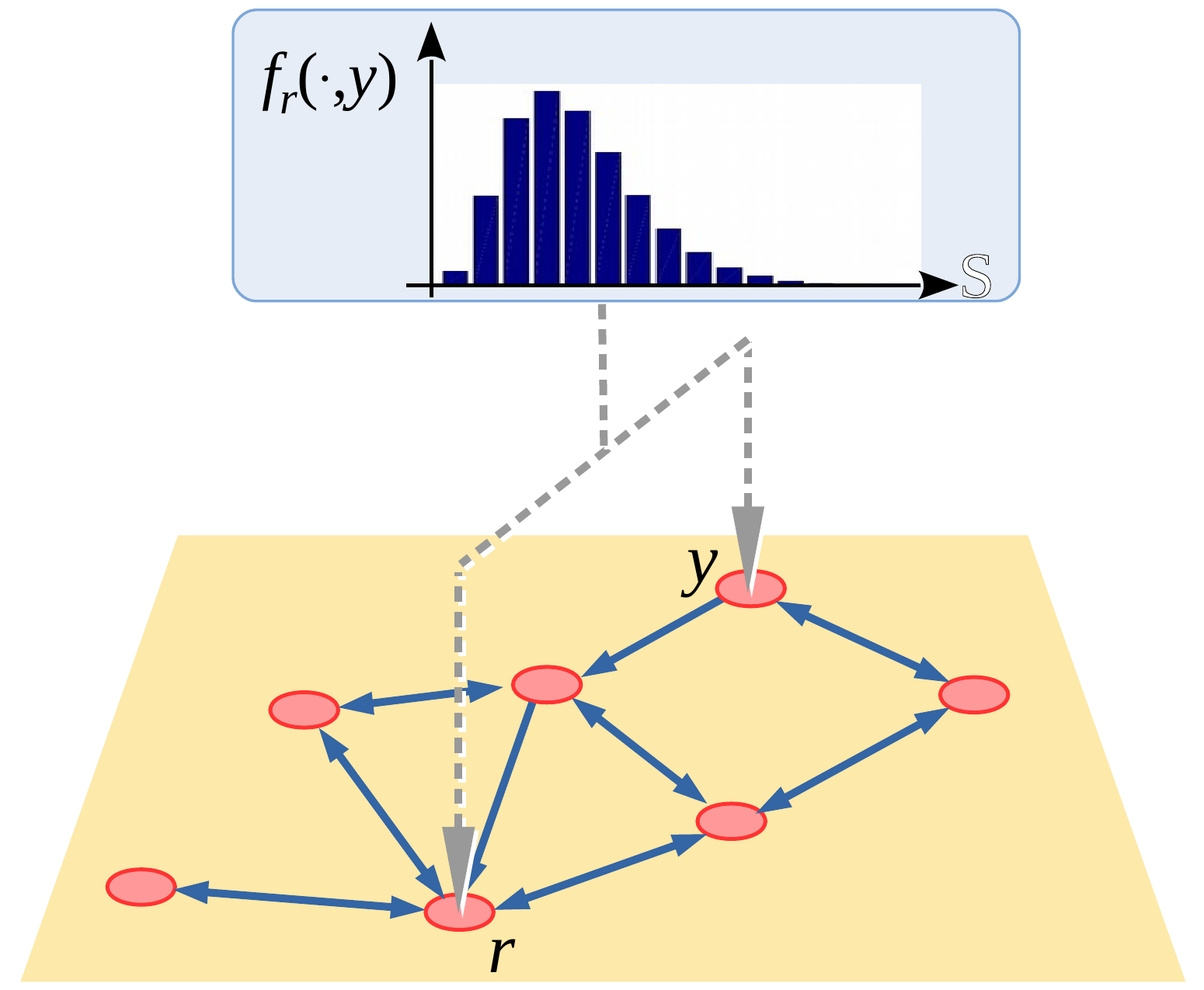} 
	\caption{\small 
		\changeD{For fixed $r,y$, the plot shows the probability distribution over the signal space $\mathbb{S}$.}
	}\vspace{-0.15in} 
	\label{fig:e_of_f}
\end{wrapfigure}

\mysubsubsection{Sensor Model}
We assume that 
the probability that a pursuer at $r \in V$ measures signal $s$ (\changed{in some discrete signal space $\mathbb{S}$}) using its on-board sensors \changeJ{if the} evader is at $y\in V$ is given by the \emph{probability distribution} $f_{r}:\mathbb{S}\times V \rightarrow \mathbb{R}_+$,
$f_{r}(s,\evaderpos) ~~=~~ \mathbb{P}(\mathcal{S} = s   \mid  \mathcal{Y} = \evaderpos)$
%
where, $\mathcal{S}$ is 
the random variable for signal measurement, and
$\mathcal{Y}$ is the random variable for evader position.
Likewise, $h_y(s,r) = \mathbb{P}(\mathcal{S} = s   \mid  \mathcal{R} = r)$ is the senor model used by the evaders giving the probability that an evader at $y$ measures signal $s$ when a robot is at $r$.

%
Using Bayes' rule, \changed{the updated probability distribution of the $j^\text{th}$ evader as computed by a pursuer at, $r$, based on sensor \emph{measurement}, $s^t$, and the prior probability estimate, $\widetilde{p}_j^t$, is} 
\vspace{\eqnpregap}
{\eqnsize \begin{eqnarray} \label{eq:Y-given-S-expanded}
p_j^t(y) & = & \mathbb{P}(\mathcal{Y}_j = y \mid \mathcal{S}_j = s^t) \nonumber \\
	& = & \mathbb{P}(\mathcal{S}_j = s^t \mid  \mathcal{Y}_j = \evaderpos)~ \frac{\mathbb{P}(\mathcal{Y}_j = \evaderpos)}{\mathbb{P}(\mathcal{S}_j = s^t)} \nonumber
\end{eqnarray}

\vspace{-1.3em}
\begin{eqnarray}
	& = & \quad \frac{f_{r}(s^t,y) ~ \widetilde{p}_j^{t}(y)}{\sum_{\evaderpos'\in V} f_{r}(s^t,\evaderpos') ~\widetilde{p}_j^t(y')} 
	\end{eqnarray}}

\noindent
If multiple signals, $s_1^t, s_2^t, \cdots$, are received by robots $r_1, r_2, \cdots$ at a time step, they are incorporated 
in sequence:

\vspace{1.0\eqnpregap}
{\eqnsize \begin{equation}
	p_j^t(y) = 
	\prod_{l} \frac{f_{r_l}(s_l^t,\evaderpos)}{\sum_{\evaderpos'\in V} f_{r_l}(s_l^t,\evaderpos') ~\widetilde{p}_j^t (y') } ~\widetilde{p}_j^t(y)
	\end{equation}
}

\vspace{1.0\eqnpregap}\noindent
Likewise, the evaders $y_1, y_2, \cdots$ measuring signals $s_1^t, s_2^t, \cdots$ update the probability distributions that they use to represent the $i^\text{th}$ pursuer according to

\vspace{1.0\eqnpregap}
{\small \begin{equation}
	q_i^t(r) = 
	\prod_{l} \frac{h_{y_l}(s_l^t,r)}{\sum_{r'\in V} h_{y_l}(s_l^t,r') ~\widetilde{q}_i^t (r') } ~\widetilde{q}_i^t(r)
	\end{equation}}

\vspace{\sectiongaps}
\mysubsection{Evader Control Strategy}

In presence of pursuers, an evader $y_j$ actively tries to move away from the pursuers targeting it.
With the evader at $y\in V$ and deterministic pursuers, $\{r_i\}_{i\in I_j}$, trying to capture it, we define a \emph{mean capture time} as follows
	
	\vspace{1.2\eqnpregap}
	{\eqnsize \begin{equation} \label{eq:aveGScore}
		\tau(y, \{r_i\}_{i\in I_j}) = \frac{1}{\displaystyle \sum_{i \in I_j} \frac{1}{\widetilde{d}_g(r_i,y) / v_i} }
		\end{equation}}
	
	\vspace{0.9\eqnpostgap}
	\noindent
where $\widetilde{d}_g(r_i,y) = \max \left(0, ~ d_g(r_i,y) - \rho_i \right)$ is the \emph{effective geodesic distance}
	between $r_i,y\in V$, which accounts for the fact that robot $r_i$ has a capture radius of $\rho_i$.
%
As described above, $\tau(y, \{r_i\}_{i\in I_j})$ is the harmonic mean of the \emph{effective least time} to be taken by the \changeD{pursuers} in $I_j$ to reach the point $y$.
	$\tau$ is thus a function that has higher value on the vertices in $V$ that are farther away from the pursuers in $I_j$.
%
	The reason behind taking harmonic mean 
	is that the harmonic mean gets lower contribution from distant pursuers and higher contribution from the nearby pursuers. 

In order to determine the best action that the evader at $y' \in V$ can take, 
it computes the marginal increase in $\tau$ if it moves to $y \in V$:

\vspace{1.5\eqnpregap}
{\small \begin{equation}\!\!\!\!\begin{array}{l}
\Delta\tau(y, y', \{r_i\}_{i\in I_j}) \\ \qquad\qquad =~ \max\left(0, \tau(y, \{r_i\}_{i\in I_j}) \!-\! \tau(y', \{r_i\}_{i\in I_j}) + \epsilon \right)
\end{array} \label{eq:marginal-increase-tau}\end{equation}}

\vspace{1.5\eqnpregap}\noindent
where $\epsilon$ is a small number 
that gives a small positive marginal increase for some neighboring vertices in scenarios when the evader gets cornered against an obstacle.

\vspace{-0.5\sectiongaps}
\mysubsubsection{Evader's Control Strategy}
\label{sec:evader-control}
In a deterministic setup the evader at $y'$ will move to

{\eqnsize \begin{equation}
y^*_j(y', \{r_i\}_{i\in I_j}) ~:=~ \arg\!\max_{y\in A_{y'}} \Delta\tau(y, y', \{r_i\}_{i\in I_j})
\end{equation}}

\begin{wrapfigure}{r}{0.32\columnwidth} \vspace{-0.15in} 
	\centering
	%
	%
	\includegraphics[width=0.32\columnwidth,trim=0 0 0 20,clip=true]{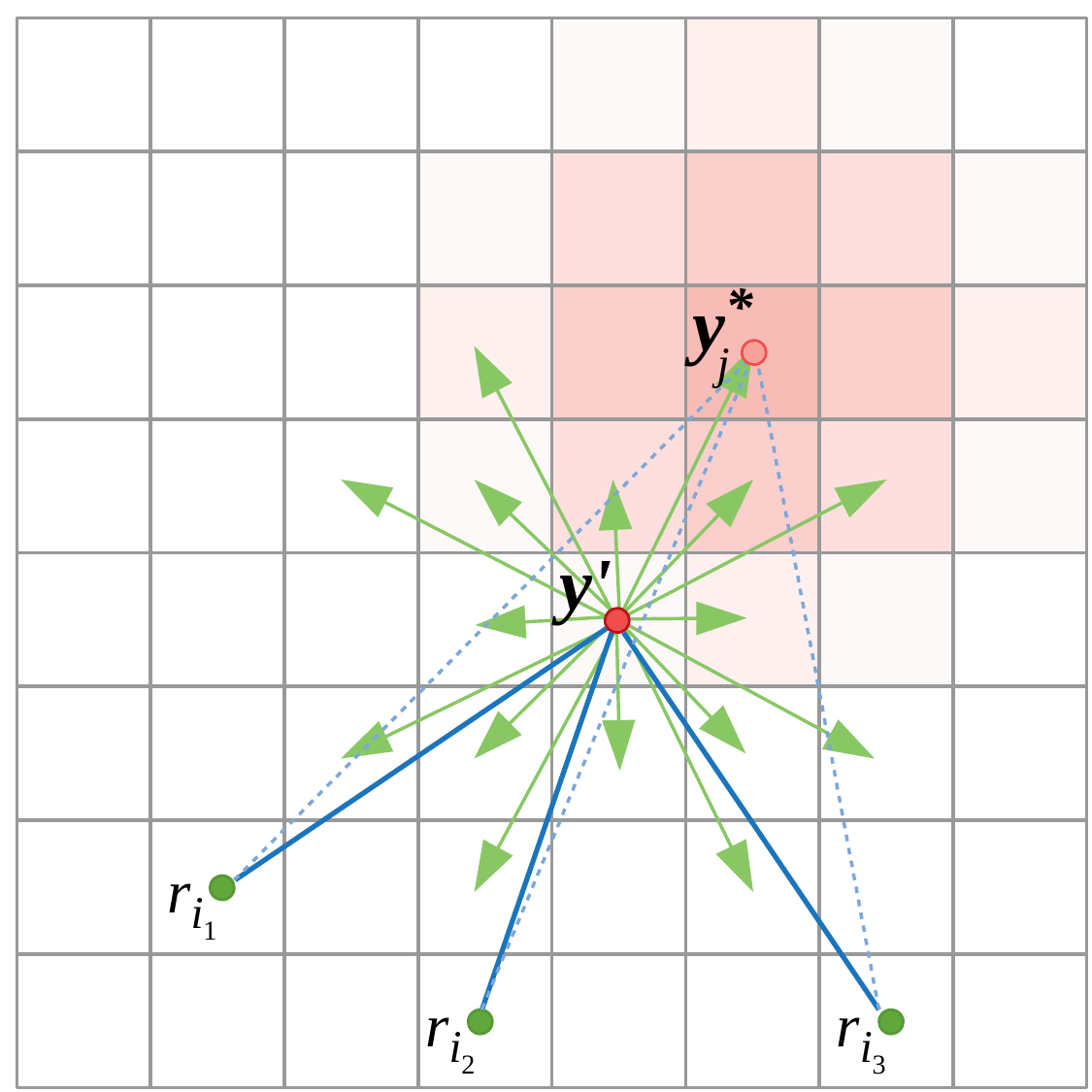} \label{fig:K-active}
	%
	\vspace{-0.2in} \caption{
		Discrete representation of an environment and \changedB{illustration of} control strategy of evader at $y'$.
		Transition probabilities, $K_j(\cdot,y')$ are shown in light red shade.
	} 
	%
	\label{fig:eight_connect_graph}
\end{wrapfigure}

\noindent 
where $A_{y'}$ refers to the states/vertices in the vicinity of $y'$ that the evader can transition to in the next time-step. But, in the probabilistic setup where the evaders represent the $i^\text{th}$ pursuer by the distribution $q_i$, with every $y\in A_{y'}$ an evader associates a probability that it is indeed the best transition to make.
In practice, these probabilities are computed by sampling $\{r_i\}_{i\in I_j}$ from the distributions $\{q_i\}_{i\in I_j}$, and counting the proportion of samples for which a $y\in A_{y'}$ is the neighbor that maximizes the marginal increase in capture time. The evader then uses this probability distribution over its neighboring states to make a stochastic transition.

\mysubsubsection{Pursuer's Prediction of Evader's Distribution}
\label{sec:pursuer-prediction-of-evader}
The pursuers know the evader's strategy of maximizing the marginal increase in capture time. However, they do not know the 
the distributions, $q_i$, that the evaders maintain of the pursuers. 
The uncertainty in the action of the evader due to that is modeled by a 
normal distribution centered at $y^*_j(y', \{r_i\}_{i\in I_j})$.
If the evader is at $y'$, the transition probability $K_j(y,y')$ is the assumed to be

{\eqnsize \begin{equation} \label{eq:K-j}
 \!\!\!\! K_j(y,y') = \left\{ \!\!\!\! \begin{array}{ll} \kappa_j \exp\left(-\frac{d_f\left(y, ~y^*_j(y', \{r_i\}_{i\in I_j})\right)^2}{2 \sigma_j^2}\right)\!\!, & \!\!\!\!\text{if $y\in A_{y'}$} \\ 0, & \!\!\!\!\text{otherwise.} \end{array} \right. \!\!\!\!
\end{equation}}

\noindent
where, for simplicity, $d_f$ is assumed to be the Euclidean distance between the neighboring vertices in the graph, and $\kappa_j$ is a normalization factor so that $\sum_{y\in V} K(y,y') = 1$.

\vspace{1em}
\mysubsection{Pursuer Control Strategy}
\label{sec:control}

\begin{wrapfigure}{r}{0.31\columnwidth} 
	\vspace{-0.3in} 
	\centering
	\fbox{\includegraphics[width=0.3\columnwidth, trim=20 0 0 0, clip=true]{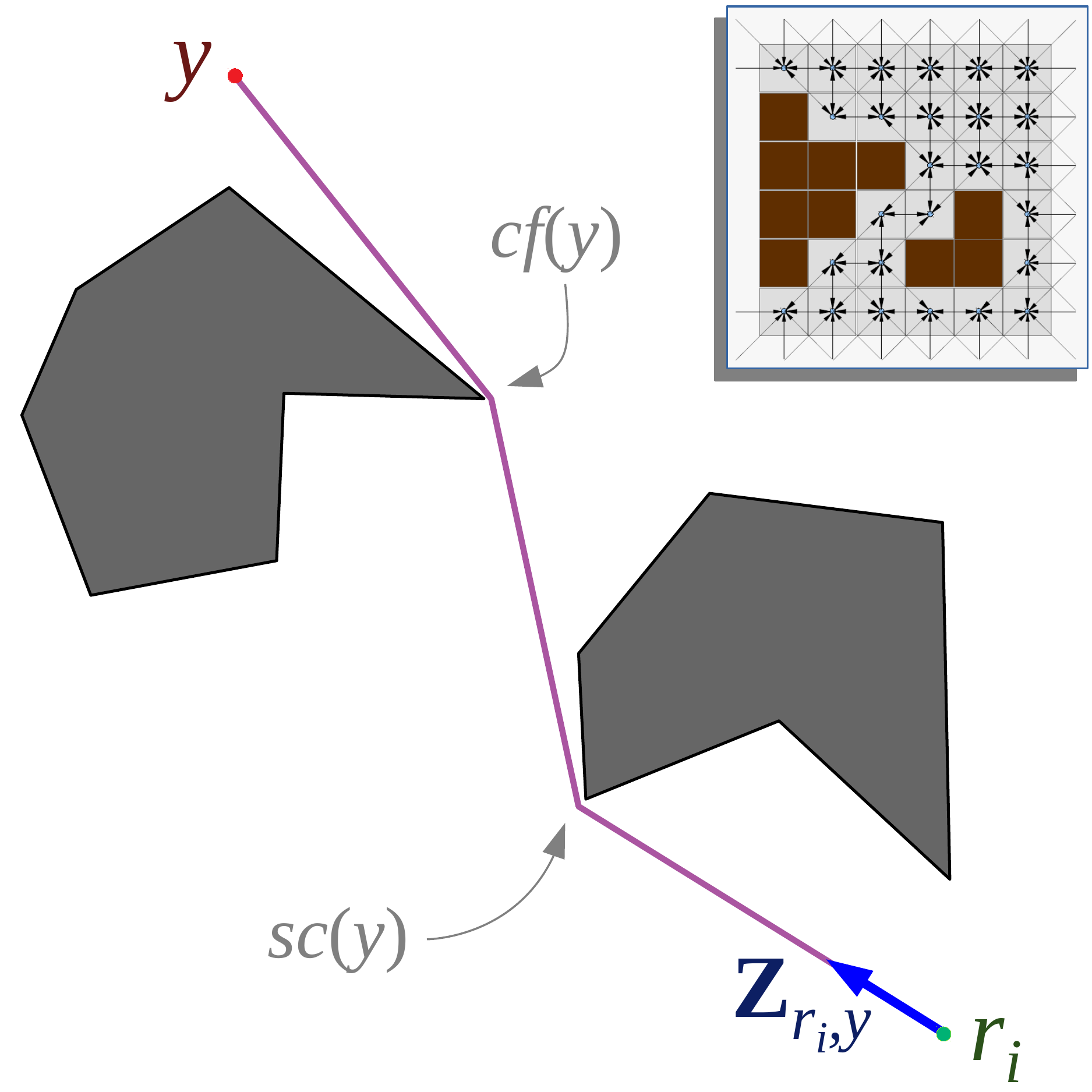}} 
	\vspace{-0.8em}\caption{\small Theta* algorithm is used on a $8$-connected grid graph, ${G_{+\hspace{-2.0mm}\times}}$ (top right inset) for computing geodesic distances as well as control velocities for the pursuers.}  \label{fig:geodesic}\vspace{-0.2in} 
\end{wrapfigure}

A pursuer, $r_i\in I_j$, pursuing the evader at $y_j$ needs to compute a velocity for doing so.
	In a deterministic setup, if the evader is at $y_j\in V$, 
	the pursuer's control strategy is to follow the shortest (geodesic) path in the environment connecting $r_i$ to $y_j$.
	This controller, in practice, can be implemented as a
	gradient-descent of the square of the \emph{path metric} (geodesic distance)
	and
	is given by $\mathbf{v}_i = k \frac{\partial d_g(r_i,y_j)^2}{\partial \mathbf{X}(r_i)} = 2k ~d_g(r_i,y_j) ~\hat{\mathbf{z}}_{r_i,y_j}$, where $k$ is a proportionality constant, $d_g(r_i,y_j)$ is the shortest path (geodesic) distance between $r_i$ and $y_j$, and $\hat{\mathbf{z}}_{r_i,y_j}$ is the unit vector to the shortest path at $r_i$ (see Figure~\ref{fig:geodesic}).
	Such a controller does not suffer from local minimas due to presence of non-convex obstacles since the geodesic paths \emph{go around} obstacles.
	A formal proof of that and the fact that $\frac{\partial d_g(r,y)}{\partial \mathbf{X}(r)} = \hat{\mathbf{z}}_{r,y}$, appeared in \cite{coverage:riemannian:IJRR:13}.).
	This gives a simple velocity controller for the pursuer.
	
\mysubsubsection{Pursuer's Control Strategy}
Since the pursuers describe the $j^\text{th}$ evader's position by the probability distribution $p_j^t$ over $V$, we compute an expectation on the velocity vectors of the $i^\text{th}$ pursuer (with $i\in I_j$) as follows:

\vspace{1.2\eqnpregap}
{\eqnsize \begin{equation} \label{eq:contol-vel}
	\mathbf{\hat{v}}_i = \sum_{y\in V} 2k ~d_g(r_i,y) ~\hat{\mathbf{z}}_{r_i,y} ~p_j^t(y)
	\end{equation}}

\vspace{0.8\eqnpostgap}\noindent
Since the pursuer has a maximum speed of $v_i$, and the exact location of the evader is unknown, we always choose the maximum as speed for the pursuer:
$\mathbf{v}_i ~=~ v_i ~\frac{\mathbf{\hat{v}}_i}{\|\mathbf{\hat{v}}_i\|}$.

For computing $d_g(r_i,y)$ we use the Theta* search algorithm~\cite{conf-aaai-NashDKF07} on a uniform $8$-connected square grid graph, $G_{+\hspace{-2.19mm}\times}$, representation of the environment.
While very similar to Dijkstra's and A*, Theta* computes paths that are not necessary restricted to the graph and are closer to the true shortest path in the environment.
Computation of the sum in equation~\eqref{eq:contol-vel} can be performed during the Theta* search.
Algorithm \ref{alg:pursuer_control} \changed{describes the computation of 
	$d_{g}(r_{i}, y)$ (the shortest path (geodesic) distance between $r_i$  and a point $y$ in the environment) and the control velocity $\mathbf{v}_i$.}
The algorithm is reminiscent of Dijkstra's search, maintaining an open list, $Q$, and expanding the least $g$-score vertex at every iteration, except that the \emph{came-from} vertex ($cf$) of a vertex can be a distant predecessor determined by line of sight (Lines~\ref{lin:los-start}--\ref{lin:los-end}) and the summation in \eqref{eq:contol-vel} is computed on-the-fly during the execution of the search (Line~\ref{ln:vel-computation}).

\input{theta_star}

\mysubsubsection{Evader's Prediction of Pursuer's Distribution}
\label{sec:evader-prediction-of-pursuer}

Since the evaders represent the $i^\text{th}$ pursuer using the probability distribution $q_i$, 
they need to predict the pursuer's probability distribution in the next time step knowing the pursuer's control strategy.
This task is assigned to the $j^\text{th}$ evader such that $i\in I_j$ (we define $\overline{j}(i)$ to be the index of the evader assigned to pursuer $i$).
It
executes Theta* algorithm, similar to Algorithm \ref{alg:pursuer_control}, but the start vertex being $y_j$. 
Once executed, 
the line segment connecting any $r'\in V$ and $cf(r')$ gives the direction in which the $i^\text{th}$ pursuer at $r'$ would tentatively move in the next time-step based on the aforesaid control strategy of the pursuer. Knowing the speed of a pursuer, the evader can thus compute the next position of the pursuer, $r^*_j(r', y_j)$, if it is currently at $r'$.
However, in order to account for the fact that the pursuer does not precisely know the evader's position (and instead use the distribution $p_j$ to represent it), analogous to \eqref{eq:K-j}, we use the following transition probability for the \emph{prediction} step of updating $q_i$

{\eqnsize \begin{equation}
L_i(r,r') = \left\{ \!\! \begin{array}{ll} \kappa_i \exp\left(-\frac{d_f\left(r, ~r^*_j(r', y_{\overline{j}(i)})\right)^2}{2 \sigma_i^2}\right), & \text{if $r\in A_{r'}$} \\ 0, & \text{otherwise.} \end{array} \right.
\end{equation}}

\noindent
where $\kappa_i$ is the normalization factor.
}

\input{assignment_new} \label{sec:assign}


\input{results}

%

\vspace{-1em}
\mysection{CONCLUSIONS}

In this paper, we considered a 
pursuit-evasion problem with multiple pursuers, and
\changed{multiple evaders under uncertainties. \changeD{Each type of agent (pursuer or evader) represents the individuals of the other type using probability distributions that they update based on known control strategies and noisy sensor measurements.} 
	Markov
localization is used to update a probability distributions. \changeD{The evaders use a control strategy to} actively evade the pursuers,
\changeD{while each pursuer use a}
control algorithm based on Theta* search for reducing the expected distance to
the probability distribution of the evader that it's pursuing. We used a novel redundant pursuer} assignment algorithm which utilizes an excess
number of pursuers to minimize the total \changeJ{or maximum} expected time to capture the
evaders. \changeJ{Our simulation results have shown a \changeC{consistent and statistically significant} reduction of
time to capture when compared against a nearest-neighbor algorithm.}

\vspace{-0.3em}
\bibliographystyle{ieeetr}
\bibliography{references,references2}



\end{document}

%% file: theta_star.tex
%
%
%
%
%
%
%
%
%

\setlength{\textfloatsep}{0pt}
{\small \begin{algorithm}[h] 
\caption{Theta* Based Pursuer Control} \label{alg:pursuer_control}

{\algsize

\textbf{Inputs:}
\textit{i.} Graph $G_{+\hspace{-2.19mm}\times}=(V,E)$;~
\textit{ii.} \changeD{Pursuer} location $r_i\in V$;~
\textit{iii.} Evader probability distribution $p_j$.

\textbf{Outputs:}
\textit{i.} The lengths of shortest paths to all vertices $g:V \rightarrow \mathbb{R}_+$;~
\textit{ii.} The control velocity $\mathbf{v}_i$.

\begin{algorithmic}[1]
\State \changed{$g(y) \leftarrow \infty$, ~$c\!f(y) = \emptyset$, ~$sc(y) = \emptyset$ ~for all $y\in V$} 
		\newline ~\algcomment{\^{}\^{}\^{} $g$-score, \emph{came-from} vertex and \emph{second-from-start} vertex.}
\State $Q=\emptyset$, ~$Z=\emptyset$ \algcomment{open and closed lists}
\State \changed{$g(r_i) \leftarrow 0$, ~$cf(r_i) \leftarrow r_i$}, ~$Q=Q \cup \{r_i\}$ \label{ln:init}
\State Set $\mathbf{\tilde{v}}_i \leftarrow \pmb{0}$
\While {(\changed{$Q \neq \emptyset$})}
	\State Set $y \leftarrow \text{argmin}_{y' \in Q} g(y')$ \algcomment{pop heap}
	\State Set \changed{$Q \leftarrow Q - \{y\}$,} ~$Z \leftarrow Z \cup \{y\}$
	\ForEach {($\{ w \in  \mathcal{N}_{G_{+\hspace{-1.85mm}\times}}(y) \mid w \notin Z\}$)} \label{ln:neighbors}
        \changed{
		\If {~~(LineOfSight$(cf(y), w)$)} \label{lin:los-start}
            \State Set $\overline{cf} \leftarrow cf(y)$ \algcomment{potential came-from vertex}
        \Else
            \State Set $\overline{cf} \leftarrow y$ \algcomment{potential came-from vertex}
        \EndIf \label{lin:los-end}
        \State Set $\overline{g} \leftarrow g(\overline{cf}) + \text{d}_{E}(\overline{cf}, w)$
		\If {($\overline{g} < g(w)$)}
			\State Set $g(w) \leftarrow \overline{g}$
			\State Set $cf(w) \leftarrow  \overline{cf}$
            \If {($cf(w) ~!\!= r_i$)}
                \State Set $sc(w) \leftarrow sc(y)$
            \Else 
                \State Set $sc(w) \leftarrow w$
            \EndIf
		\EndIf
    }
	\EndFor
    \changed{
    \If {($y ~!\!= r_i$)} \algcomment{control computation}
        \State Set $\hat{\mathbf{z}}_{r_i,y} \leftarrow \frac{\mathbf{X}(sc(y)) - \mathbf{X}(r_{i})}{||\mathbf{X}(sc(y)) - \mathbf{X}(r_{i})||}$
        \State Set $\mathbf{\tilde{v}}_i \leftarrow \mathbf{\tilde{v}}_i + 2k ~g(r_i,y) ~\hat{\mathbf{z}}_{r_i,y} ~p_j(y)$ \label{ln:vel-computation}
    \EndIf
    }
\EndWhile
\State Set $\mathbf{v}_i \leftarrow v_i ~\frac{\mathbf{\tilde{v}}_i}{\|\mathbf{\tilde{v}}_i\|}$
\end{algorithmic}}
\end{algorithm}}


We start the algorithm by initiating the \changed{\emph{open list}} with \changed{the single start vertex,} $r_i$, \changed{set its $g$-score to zero, and its came-from vertex, $c\!f$, to reference to itself (line~\ref{ln:init}).}
\changed{Every time a vertex, $y$ (one with the minimum $g$-score in the open list, maintained using a heap data structure), is expanded, Theta* checks for the possibility of updating a neighbor, $w$, from the set of neighbors, $\mathcal{N}_{G_{+\hspace{-1.85mm}\times}}(y)$, of the vertex that are not in the closed list (line~\ref{ln:neighbors}). Based on the existence of a}
direct line of sight from the came-from vertex of $y$ and \changed{the} vertex $w$, the potential new came-from vertex, $\overline{cf}$, is set to $cf(y)$  or $y$. 
\changed{The new potential $g$-score is computed as the sum of the $g$-score of $\overline{cf}$ and the Euclidean distance, $\text{d}_E(\overline{cf}, w) = \|\mathbf{X}(\overline{cf}) - \mathbf{X}(w)\|$, between the two vertices. If lower,} $g(w)$ \changed{is} updated, the came-from vertex of $w$ \changed{is set to $\overline{cf}$, and the vertex on the path second from the start, $sc(w)$, is copied from that of $y$ unless $w$ is itself second from start.
We also compute the control velocity as part of the Theta* search algorithm.
Every time a vertex is expanded, we add the corresponding term in the summation of equation~\ref{eq:contol-vel} to the vector $\mathbf{\hat{v}}_i$ (line~\ref{ln:vel-computation}), which we scale to have magnitude of the maximum possible speed of the pursuer, $v_i$, at the end.
}

%% file: assignment_new.tex
\mysection{\changedB{Pursuer-to-Evader Assignment Strategy}}
\vspace{0.2em}

The goal for our assignment strategy is to try to find the assignment that minimizes \changeJ{either} the \emph{total expected capture time} \changeD{(the sum of the times taken to capture each of the evaders in $\mathcal{C}_y$)} \changeJ{or the \emph{maximum expected capture time} \changeD{(the maximum out of the times taken to capture each of the evaders in $\mathcal{C}_y$)}}.
%
\changed{We assume that there are more pursuers in the environment than the number of evaders. Starting with an initial optimal assignment of the evaders, we determine the assignment of the remaining pursuers. To that end we use the algorithm outlined in~\cite{prorok2020robust}.}
\changeD{We first consider the assignment problem from the perspective of the pursuers -- with the evaders represented by probability distributions $\{p_j\}_{j\in \mathcal{C}_y}$, what's the best pursuer-to-evader assigment?}

\vspace{-0.6em}
\mysubsection{Probabilistic Assignment Costs}

\changed{Since the evader $j$ is represented by the probability distribution, $p_j$, over $V$,
we \changeJ{denote} $T_{ij}$ as the random variable representing the uncertain travel time from pursuer $i$ to evader $j$. That is,

\vspace{1.2\eqnpregap} 
{\eqnsize 
\[
\mathbb{P}\left( T_{ij} \in [\tau, \tau+\Delta\tau) \right) ~~=\!\!\!\!\! \displaystyle \sum_{ \{ y~\in~ V ~| \atop \frac{1}{v_i} d_g(r_i,y) \in [\tau, \tau+\Delta\tau) \} } \!\!\!\! p_j(y) \] 
}

\vspace{\eqnpostgap} \noindent
We first note that $T_{ij}$ and $T_{i'j'}$ are independent variables whenever $j$ and $j'$ are different (\emph{i.e.}, the time taken to reach evader $j$ does not depend on time taken to reach evader $j'$).
However, $T_{ij}$ and $T_{i'j}$ are dependent random variables since, for a given travel time (and hence travel distance) from pursuer $i$ to evader $j$, and knowing the distance between pursuers $i$ and $i'$, the possible values of distances between pursuer $i'$ and evader $j$ are constrained by \changedB{the} triangle inequality.
That is, for any given $j$, the random variables in the set $\{T_{ij} ~|~ i\in \mathcal{I}\}$, where $\mathcal{I}$ is a set of pursuer indices, are dependent.
This can be seen more clearly by considering a potential evader position $y\in V$ which has an associated probability of $p_j(y)$. Given that position, $\frac{1}{v_i} d_g(y,r_i)$ 
is the time taken by the pursuer $i\in\mathcal{I}$ to reach the evader.
In particular, the following holds:

\vspace{2\eqnpregap} 
{\eqnsize
\begin{eqnarray}
\!\!\!\!\!\!\!\!& & \!\!\!\!\!\!\!\! 
\mathbb{P}\left( \bigwedge_{i\in\mathcal{I}} ~T_{ij} \in [\tau_i, \tau_i+\Delta\tau_i) \right) \nonumber \\
& = & \changeD{\sum_{y} \mathbb{P}\left( \bigwedge_{i\in\mathcal{I}} \frac{1}{v_i} d_g(r_i, y) \in [\tau_i, \tau_i+\Delta\tau_i) \right)} \nonumber \\
& = & \!\!\!\!\!\!\!\! \changeD{\sum_{\{y \in V ~|~ \atop \qquad\frac{d_g(r_i, y)}{v_i} \,\in\, [\tau_i, \tau_i+\Delta\tau_i), \forall i\in\mathcal{I} \}} \!\!\!\!\!\!\!\!\!\!\!\! p_j(y) }
%
\label{eq:sampling}
\end{eqnarray}
}

\vspace{0.8\eqnpostgap}\noindent
Thus, in order to compute the \changeD{joint probability} distributions of $\{T_{ij} ~|~ i\in \mathcal{I}\}$,
we can sample a $y$ from
\changeD{the probability distribution $p_j$}
and compute the travel times $\tau_i = \frac{1}{v_i} d_g(r_i, y), ~i\in \mathcal{I}$, \changeD{and hence populate the distribution. 
}
}

\mysubsection{Expected Capture Time Minimization For an Initial One-to-One Assignment}

\changed{In order to determine an \emph{initial} assignment $\mathcal{A}_0 \subseteq \mathcal{F}$ such that exactly one pursuer is assigned to an evader (thus potentially allowing unassigned pursuers).
Since
for every $(i,j),(i',j')\in \mathcal{A}_0$, $T_{ij}$ and $T_{i'j'}$ are independent variables,
the problem of finding the optimal initial assignment that 

\noindent
minimizes the \emph{total expected capture time} becomes\footnote{The expectation of the sum of two or more independent random variables is the sum of the expectations of the variables. 
}

~

{\eqnsize \begin{eqnarray} \vspace{-2em}
 \!\!\!\!   \mathcal{A}_0 & = &  \!\!\!\!\argmin_{\substack{\mathcal{A}'\subset \mathcal{F} ~~\text{s.t.} \\ (i,j),(i',j')\in \mathcal{A}' ~\Rightarrow ~ i\neq i', j\neq j'} } \mathbb{E}\left( \sum_{(i,j)\in\mathcal{A}'} T_{ij} \right) \nonumber \\
  \!\!\!\!       & = &  \!\!\!\!\argmin_{\substack{\mathcal{A}'\subset \mathcal{F} ~~\text{s.t.} \\ (i,j),(i',j')\in \mathcal{A}' ~\Rightarrow ~ i\neq i', j\neq j'} } ~~\sum_{(i,j)\in\mathcal{A}'} \mathbb{E}\left( T_{ij} \right)  ~~
\end{eqnarray}}

\vspace{-1em}
Thus, for computing the initial assignment, it is sufficient to use the numerical costs of $C_{ij} = \mathbb{E}\left( T_{ij} \right) $ in the assignment of pursuer $i$ to evader $j$, and thus find an assignment that minimizes the net cost. 
In practice we use a Hungarian algorithm to compute the assignment. \changeJ{While a Hungarian algorithm is an efficient method for computing the assignment that minimizes the expected \emph{total} time of capture, 
	generalizing it to the problem of minimizing the expected \emph{maximum} capture time is non-trivial, which we address next.
}}

\subsubsection{Modified Hungarian Algorithm for Minimization of Maximum Capture Time}\label{sec:modified-hungarian}

\changeJ{For finding the initial assignment that minimizes the \emph{maximum expected capture time}, we develop a modified version of the Hungarian algorithm. To that end we observe that in a Hungarian algorithm, instead of using the the expected capture times as the costs, we can use the $p$-th powers of the expected capture times, $C_{ij} = \left(\mathbb{E}\left( T_{ij} \right)\right)^p$. Making $p \rightarrow \infty$ results in the appropriate cost that makes the Hungarian algorithm compute an assignment that minimize the maximum expected capture time (the infinity norm). However, for computation we cannot practically raise a number to infinity, and thus need to modify the Hungarian algorithm at a more fundamental level.
	
In a simple implementation of the Hungarian algorithm~\cite{munkres1957algorithms}, one performs multiple row and column operations on the cost matrix wherein
a specific element of the cost matrix, $C_{i'j'}$, 
is added or subtracted from all the elements of a selected subset of rows and columns. 
Thus, if we want to use the $p$-th powers of the costs, but choose to maintain only the costs in the matrix (without explicitly raising them to the power of $p$ during storage), for the row/column operations we can simply raise the elements of the matrix to the power of $p$ right before the addition/subtraction operations, and then take the $p$-th roots of the results before updating the matrix entries.
That is, addition of $C_{i'j'}$ to an element $C_{ij}$ will be replaced by the operation $C_{ij} \oplus_p C_{i'j'} =  \sqrt[p]{C_{ij}^p + C_{i'j'}^p}$, and subtraction will be replaced by the operation $C_{ij} \ominus_p C_{i'j'} = \sqrt[p]{C_{ij}^p - C_{i'j'}^p}$.
Thus, letting $p \rightarrow \infty$, 
we have $C_{ij} \oplus_\infty C_{i'j'} = \max\{C_{ij}, C_{i'j'}\}$
and $C_{ij} \ominus_\infty C_{i'j'} = {\small \begin{cases}
C_{ij},& C_{ij} > C_{i'j'} \\
0,& C_{ij} = C_{i'j'} 
\end{cases}}$.
Thus, we can compute the assignment that achieves the minimization of the maximum expected capture time using this modified algorithm, but without actually needing to explicitly raise the costs to the power of a large $p \rightarrow \infty$.
}

\mysubsection{Redundant Pursuer Assignment Approach}

\changed{After computation of an initial assignment, $\mathcal{A}_0$, we determine the assignment of the remaining pursuers using the method proposed in~\cite{prorok2020robust}.}
\amanda{
Formally, we first consider the problem of selecting a set of redundant pursuer-evader matchings, $\bar{\mathcal{A}}$, that minimizes the \changed{total expected travel time} to evaders, 
under the constraint that any pursuer is only assigned once:

~

{\eqnsize \begin{eqnarray}\label{eq:supermodular}
 \!\!\!\!   \overline{\mathcal{A}} & = &  \!\!\!\!\argmin_{\substack{\mathcal{A}' \subset \mathcal{F} ~~\text{s.t.} \\ (i,j),(i',j')\in \mathcal{A}'\cup \mathcal{A}_0 ~\Rightarrow ~ i\neq i'} } ~~ \sum_{(i,j)\in\mathcal{A}'} \mathbb{E} (T_{ij}). 
\end{eqnarray}}
}

\vspace{0.6\eqnpostgap}\noindent
\amanda{Notably, the work in~\cite{prorok2020robust} shows that a cost function such
%
as~\eqref{eq:supermodular}, which considers redundant assignment under uncertain travel time, is supermodular. It follows that the assignment procedure can be implemented with a greedy algorithm that selects redundant pursuers near-optimally.}~\footnote{\amanda{We note that without an initial assignment $\mathcal{A}_0$, any solution that is smaller in size than $|\mathcal{C}_y|$ would lead to an infinite capture time, and hence, the cost function looses its supermodular property. Hence, the assumption that we already have an initial assignment is necessary.}} 

Algorithm \ref{alg:redundant} \amanda{summarizes our greedy} redundant assignment algorithm. 
\changed{At the beginning of the} algorithm, we sample $h$ \amanda{$|\mathcal{C}_r| \times |\mathcal{C}_y|$-dimensional } \changed{points from the joint probability distribution of $\{T_{ij}\}_{i\in\mathcal{C}_r,j\in\mathcal{C}_y}$ and store them in the} set $\widetilde{T}$.
\changed{In practice, the sampling is performed by sampling points, $y_j\in V$, from the evaders' probability distributions, $p_j$, for all $j\in C_y$. The travel times, $\tau_{ij} = \frac{1}{v_i} d_g(r_i,y_j), ~i\in \mathcal{C}_r, j\in \mathcal{C}_y$ then give the sample from the joint probability distributions of $\{T_{ij}\}_{i\in\mathcal{C}_r,j\in\mathcal{C}_y}$ due to equation~\eqref{eq:sampling}.
The $z^\text{th}$ sample is thus a set of travel times between every pursuer-evader pair, and will be referred to as $\widetilde{T}^z = \{\tau_{ij}^z\}_{i\in\mathcal{C}_r,j\in\mathcal{C}_y} \in \widetilde{T}$.}

\setlength{\textfloatsep}{0pt}
{\small \begin{algorithm}[h]
\caption{Total Time minimization Redundant Robot Assignment (TTRRA)}\label{alg:redundant} 

{\algsize
\textbf{Inputs:}
\textit{i.} Initial assignment, $\mathcal{A}_0$;~
\textit{ii.} $h$ samples, $\widetilde{T} = \{ \{\tau_{ij}^z\}_{i\in\mathcal{C}_r,j\in\mathcal{C}_y} \}_{z=1,2,\cdots,h}$, from the joint probability distribution of the travel times, $\{T_{ij}\}_{i\in\mathcal{C}_r,j\in\mathcal{C}_y}$.

\textbf{Outputs:}
\textit{i.} Assignment of the redundant robots, $\overline{\mathcal{A}}$;

\begin{algorithmic}[1]
\State $\overline{\mathcal{A}} \leftarrow \emptyset$
%

\For {$z \in \{1,2,\cdots,h\}$}
    \For {$(i,j)\in\mathcal{A}_0$} \label{ln:ij-loop} 
        \State $S^z_j \leftarrow \tau_{ij}^z $ \label{ln:ij-S-init}
    \EndFor
\EndFor

\For {$k \in \{1,\cdots, \changed{|\mathcal{C}_r| \!-\! |\mathcal{C}_y|} \}$}
	\State \changeJ{$T_\text{curr}^{\star} \leftarrow -\infty $}
	\State \changeJ{$T_\text{new}^{\star} \leftarrow \infty $}
	\For {$\!\{ (i,\!j) \!\in\! \changed{\mathcal{F} \!\!-\!\! \mathcal{A}_0 \!\!-\!\! \overline{\mathcal{A}} ~\big|~ 
            \mathcal{A}_0 \!\cup\! \overline{\mathcal{A}} \!\cup\! \{(i,j)\} \!\in\! \mathfrak{A} \} } \!\!$} \label{ln:loop-unmatched}
		\State $T_\text{curr} \leftarrow \frac{1}{h} \sum^{h}_{z=1} \changed{S_j^z}$
		\State $T_\text{new} \leftarrow \frac{1}{h} \sum^{h}_{z=1} \min(S_j^z, \changed{\tau_{ij}^z})$ 
		 \If  {\changeJ{$T_\text{curr} - T_\text{new} > T_\text{curr}^{\star} - T_\text{new}^{\star}$}} \label{ln:if_greater}
		 	\State \changeJ{$T_\text{curr}^{\star} = T_\text{curr}$}
		 	\State \changeJ{$T_\text{new}^{\star} = T_\text{new}$}
		 	\State $(i^{\star}, j^{\star}) \leftarrow (i, j)$
		 \EndIf
	\EndFor
	\State $\overline{\mathcal{A}} \leftarrow \overline{\mathcal{A}} \cup \{(i^{\star}, j^{\star})\}$ \label{ln:insert-assignment}
	\State $S_{j^{\star}}^z \leftarrow  \min (S_{j^{\star}}^z, \changed{\tau_{i^{\star}j^{\star}}^z}) ~~\forall z \in 1,2,\cdots,h$ 
\EndFor
\State return $\overline{\mathcal{A}}$

\end{algorithmic}}
\end{algorithm}}

\changed{In this algorithm, we first consider the initial assignment, $\mathcal{A}_0$, and collect all the sampled costs of edges incident on to the $j^\text{th}$ evader into the variable $S$. Note that a given $j\in \mathcal{C}_y$ appears in exactly one element of $\mathcal{A}_0$, thus the assignment in Line~\ref{ln:ij-S-init} assigns a value to a $S_j^z$ exactly once.
The set $\overline{\mathcal{A}}$ contains the assignment of the remaining/redundant pursuers, that we initiate with the empty set.
In Line~\ref{ln:loop-unmatched}, we loop over all the possible pursuer-to-evader pairings, $(i,j)$, that are not already present in $\mathcal{A}_0$ or $\overline{\mathcal{A}}$, and which, along with $\mathcal{A}_0$ or $\overline{\mathcal{A}}$, constitute a valid assignment.}
We go through all \changed{such potential} pairings, $(i,j)$, and pick the one with the highest marginal gain, $T_\text{curr}^{\star} - T_\text{new}^{\star}$.
The pair \changed{with} the \changed{highest} marginal gain, \changed{is thus added to $\overline{\mathcal{A}}$.}
\changed{This process is carried out $|\mathcal{C}_r|\!-\!|\mathcal{C}_y|$ times, thus ensuring that}
all pursuers \changed{get} assigned.

\changeC{
\subsubsection*{Equality in Marginal Gain}

One way that the inequality condition in Line \ref{ln:if_greater} gets violated is when the marginal gains $T_\text{curr} - T_\text{new}$ and $T_\text{curr}^{\star} - T_\text{new}^{\star}$ are equal. This can in fact happen quite often when one or more redundant robots are left to be assigned and all of them are far from all the evaders, 
rendering marginal gains for any of the assignments close to zero. 
In that case a pursuer $i$ gets randomly assigned to an evader $j$ based on the order in which the pairs $(i,j)\in \mathcal{F} \!-\! \mathcal{A}_0 \!-\! \overline{\mathcal{A}}$ are encountered in the \textbf{for} loop of Line~\ref{ln:loop-unmatched}.
In order to address this issue properly, we maintain a list of ``\emph{potential assignments}'' that corresponds to $(i,j)$ pairs (along with the corresponding $T_\text{new}$ values maintained as an associative list, $\mathcal{P\!A}^{\star}$) that produce the same highest marginal gains (\emph{i.e.}, in line~\ref{ln:if_greater} equality holds), and choose the one with the median $T_\text{new}$ value for inserting into the assignment set in Line~\ref{ln:insert-assignment}.}

\subsubsection{Redundant Robot Assignment for Minimization of Maximum Capture Time} \label{sec:MTRRA}

As for the minimization of the \emph{maximum expected capture time} in the redundant assignment process, we take a similar approach as in Section~\ref{sec:modified-hungarian}. 
We first note that choosing $(\mathbb{E}(T_{ij}))^p$ instead of simply the expected capture time in \eqref{eq:supermodular} still keep the cost function supermodular.
If we want to minimize the total (sum) expected $p$-th power of the capture time, the condition in the \emph{if} statement in line \ref{ln:if_greater} of the above algorithm needs to be simply changed to $T_\text{curr}^p - T_\text{new}^p > (T_\text{curr}^{\star})^p - (T_\text{new}^{\star})^p$.
With $p \to \infty$, this condition translates to $\max{(T_\text{curr}, T_\text{new}^{\star})} > \max{(T_\text{curr}^{\star}, T_\text{new})}$.
\changeC{Furthermore, 
to deal with the equality situations in Line \ref{ln:if_greater},
instead of choosing the assignment with the median $T_\text{new}$ from $\mathcal{P\!A}^{\star}$, we choose the one with the maximum $T_\text{new}$,
thus assigning a redundant pursuer to an evader (out of the assignments that produce the same marginal gain) that has the maximum expected capture time, thus providing some extra help with catching the pursuer.}

With these modifications, an assignment for the redundant robots can be found that minimizes the maximum expected capture time instead of total expected capture time. We call this redundant robot assignment algorithm ``\emph{Maximum Time minimization Redundant Robot Assignment}'' (MTRRA).

\changeD{ 
\subsubsection{Evader's Estimation of Pursuer Assignment}
Knowing the assignment strategy used by the pursuers, but the pursuers represented by the probability distributions $\{q_i\}_{i\in\mathcal{C}_r}$, the evaders use the exact same assignment algorithm to estimate which pursuer is being assigned to it. The only difference is that in Algorithm~\ref{alg:redundant} the elements in the input, $\tilde{T}$, are sample travel times that are computed by sampling points, $r_i$, from the probability distribution, $q_i$, for all $i\in \mathcal{C}_r$, and then computing $\tau_{ij} = \frac{1}{v_i} d_g(r_i,y_j)$ as before.
The assignment thus estimated is used by the evaders in computing their control as well as for updating the pursuers' distributions, $\{q_i\}_{i\in\mathcal{C}_r}$, as described in Sections~\ref{sec:evader-control} and \ref{sec:evader-prediction-of-pursuer} respectively.
}

%% file: results.tex

\mysection{Results} \label{sec:res}

\begin{figure} \vspace{-0.07in}
	\centering
	\subfloat[\changeJ{Max capture time in `AR0414SR' with 7 pursuers and 5 evaders.}]
	{\includegraphics[width=0.25\textwidth, trim=10 5 46 40, clip=true]{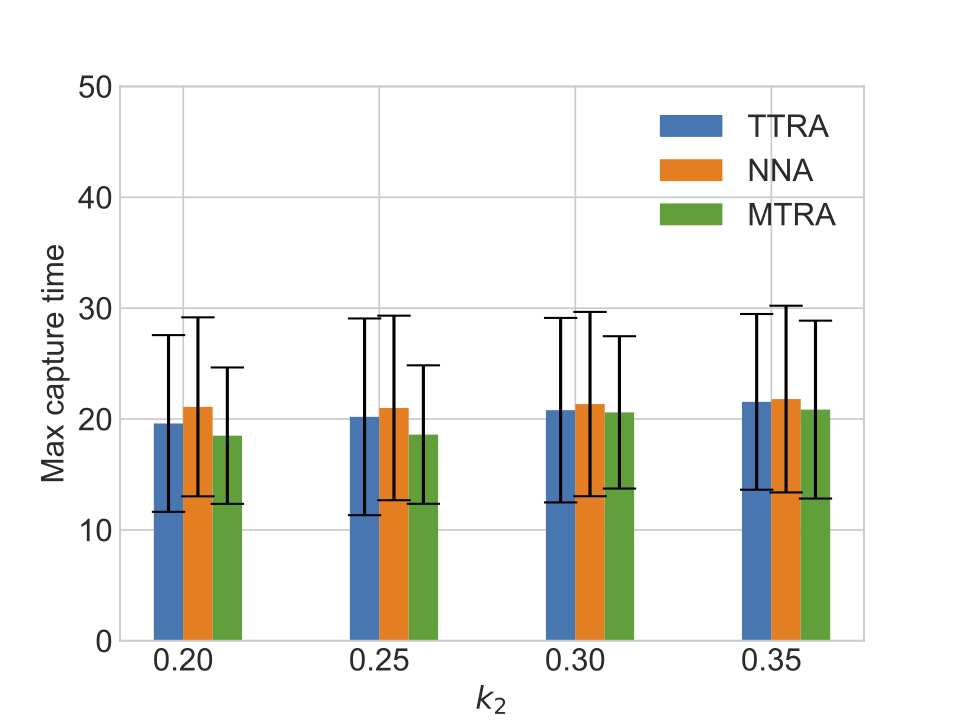} \label{fig:max_time_noise_varying}} 
	\subfloat[\changeJ{Total capture time in `AR0414SR' with 7 pursuers and 5 evaders.}]
	{\includegraphics[width=0.25\textwidth, trim=10 5 46 40, clip=true]{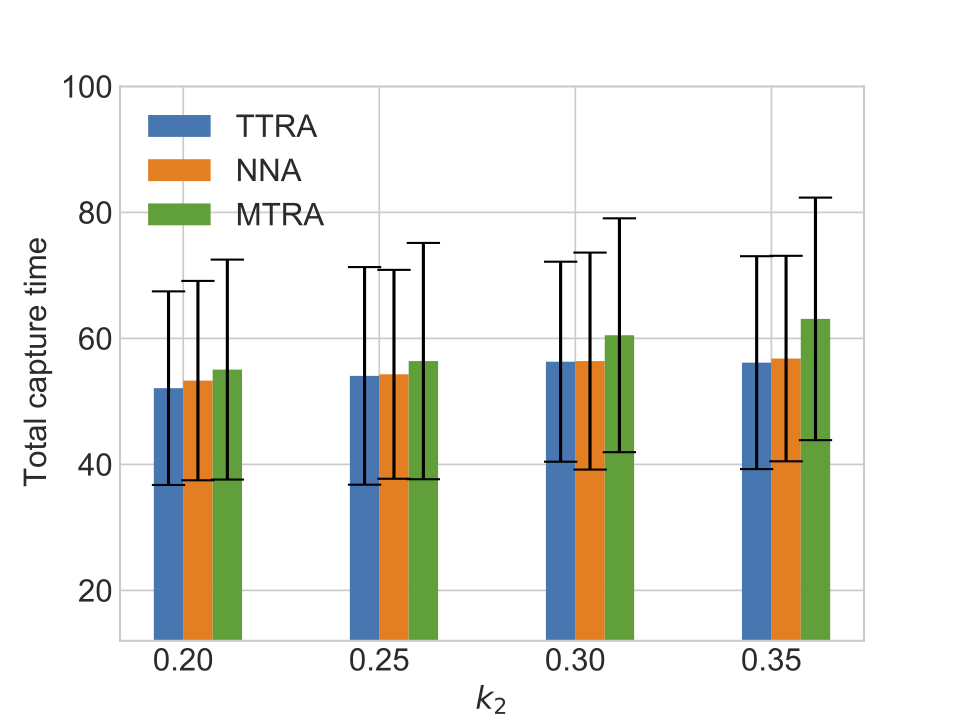} \label{fig:total_time_noise_varying}} 
	\vspace{-0.04in} \caption{\small \changeJ{The effect of varying measurement noise level on total and maximum capture time.}}
	 \vspace{0.02in} 
	\label{fig:varying_noise}
\end{figure}

\changed{\changeD{For} the} sensor models, \changeD{$f,h$}, we \changed{emulate sensing} electromagnetic \changed{radiation in the infrared or radio spectrum emitted by the \changeD{evaders/pursuers}.}
Wi-fi signals \changed{and thermal signatures \changedB{are such examples}. \changedB{For simplicity}, we ignore reflection of the radiation from surfaces, and only consider a simplified model for transmitted radiation}. 
\changed{If $I_{r,y}$ is the line segment connecting the source, $y$, of the radiation to the location of a sensor, $r$, and is parameterized by segment length, $l$, we define}
%
\emph{effective signal distance}, 
$d_{\text{eff}}(r, y) =  \int_{I_{r,y}} \!\rho(l) ~dl$,
\changed{where $\rho(l)=1$ in obstacle-free space, and $\rho_{\text{obs}}>1$ inside obstacles to emulate higher absorption of the signal.}
%
\changed{The signal space, $\mathbb{S}=\mathbb{R}_{+}$, is the space of intensity of the measured radiation, and}
\changeD{$f_{r}$ \& $h_y$ are} \changed{normal} distributions over $\mathbb{S}$ with mean $\frac{k_1}{d_{\text{eff}}(r,y)}$ and \changed{standard deviation} $\sigma = k_2 d_{\text{eff}}(r,y)$ \changed{to emulate inverse decay of signal strength and higher noise/error for larger separation (we truncate the normal distribution at zero to eliminate negative signal values).}
\changed{In all our experiments we chose $\rho_{\text{obs}} = 3, k_1 = 10$.} \changeJ{We also fix $k_2 = 0.3$, except in the experiments in Figure~\ref{fig:varying_noise}, where we evaluate the performance with varying noise level (varying $k_{2}$).}

The motion \changeD{models} for \changeD{predicting} the probability distributions are chosen as described in
\changeD{Section~\ref{sec:pursuer-prediction-of-evader} and \ref{sec:evader-prediction-of-pursuer}.
For the parameter we choose $\epsilon(y)\in(0,0.3)$ (in equation~\eqref{eq:marginal-increase-tau}) depending on whether or not $y$ is close to an obstacle. The pursuer (resp. evader) choose $\sigma_j= 0.3$ (resp. $\sigma_i= 0.3$) 
for modeling the uncertainties in the evaders' (resp. pursuers') estimate of the pursers' (resp. evaders') positions.}

\changeJ{We compared the performance of the following algorithms
	
\begin{itemize} 
	\item \emph{Total Time minimizing Robot Assignment (TTRA)}: This assignment algorithm uses the basic Hungarian algorithm for computing the initial assignment $\mathcal{A}_0$, and uses the TTRRA algorithm (Algorithm 2) for the assignment of the redundant robots at every time step. Thus the algorithm seeks to minimize the \emph{total expected capture time} (\emph{i.e.}, sum of the times to capture each evader).
	\item \emph{Maximum Time minimizing Robot Assignment (MTRA)}: This assignment algorithm uses the modified Hungarian algorithm described in Section~\ref{sec:modified-hungarian} for computing the initial assignment $\mathcal{A}_0$, and uses the MTRRA algorithm (Section~\ref{sec:MTRRA}) for the assignment of the redundant robots at every time step. Thus the algorithm seeks to minimize the \emph{maximum expected capture time} (\emph{i.e.} time to capture the last evader).
	\item \emph{Nearest Neighbor Assignment (NNA)}: In this algorithm we first construct a $|\mathcal{C}_r|\times |\mathcal{C}_y|$ matrix of expected pursuer-to-evader capture times. An assignment is made corresponding to the smallest element of the matrix, and the corresponding row and column are deleted. This process is repeated until each evader gets a pursuer assigned to it. Then we start the process all over again 
	with the unassigned pursuers and all the evaders, and the
	process continues until all the pursuers are assigned.
\end{itemize}}

\begin{wrapfigure}{r}{0.44\columnwidth} \vspace{-1.8em} 
	\centering
	\includegraphics[height=0.16\columnwidth, clip=true]{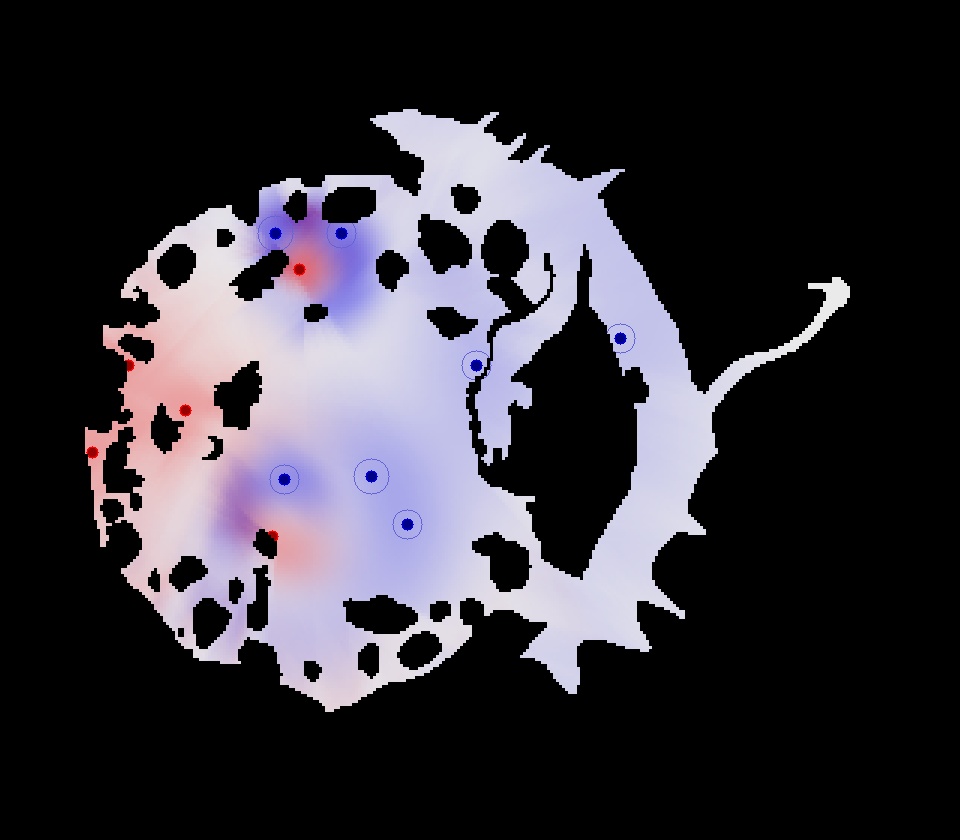} \hspace{0.1em}
	\includegraphics[height=0.16\columnwidth, clip=true]{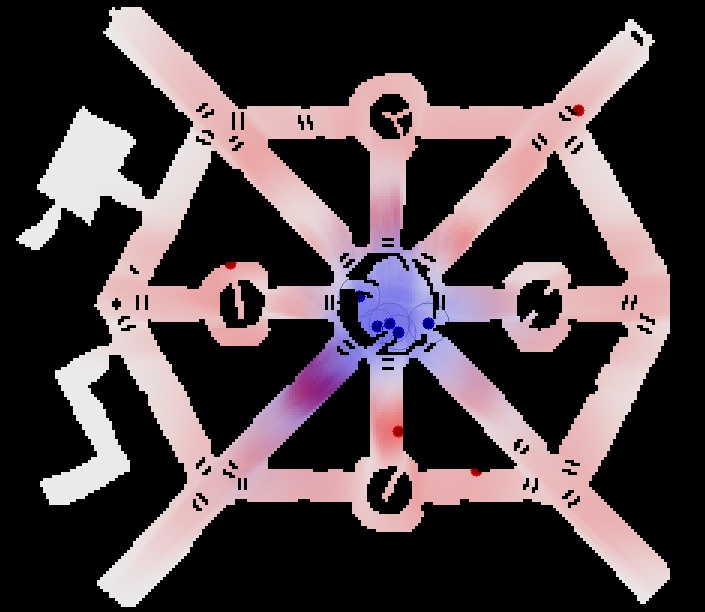}
	\vspace{-0.5em} \caption{Environments for which statistic are presented. \emph{Left:} `AR0414SR'; \emph{Right:} `AR0701SR'. See accompanying video for example simulation.} \label{fig:envs} 
\end{wrapfigure}

\noindent
We evaluated the algorithms 
 in two different environments: Game maps `AR0414SR' and `AR0701SR' from 2D Pathfinding Benchmarks~\cite{sturtevant2012benchmarks}.
%
%
\changed{For different pursuer-to-evader ratios in these environments, we ran $100$ simulations each.}
For each simulation, in environment `AR0414SR', the initial positions of pursuers and evaders \changed{were} randomly generated, while in environment `AR0701SR' the initial position of the pursuers were randomly generated in the small central circular region and the initial position of the evaders were randomly generated in the rest of the environment.
For each generated initial conditions we ran the three algorithms, TTRA, MTRA and NNA, to compare their performance.
%

\begin{figure*} 
	\centering
	\subfloat[\changeJ{Max capture time in 'AR0414SR'}]
	{\includegraphics[width=0.245\textwidth, trim=20 5 46 40, clip=true]{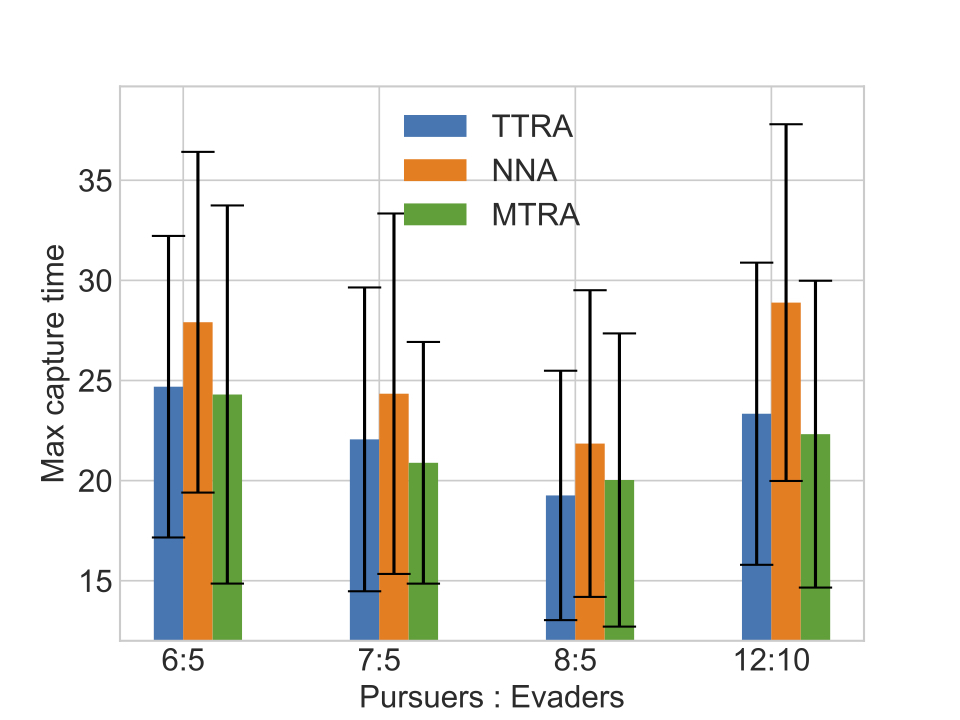} \label{fig:comparison-max-AR0414SR}} 
	\subfloat[\changeJ{Max capture time in 'AR0701SR'}]
	{{\includegraphics[width=0.245\textwidth, trim=20 5 46 40, clip=true]{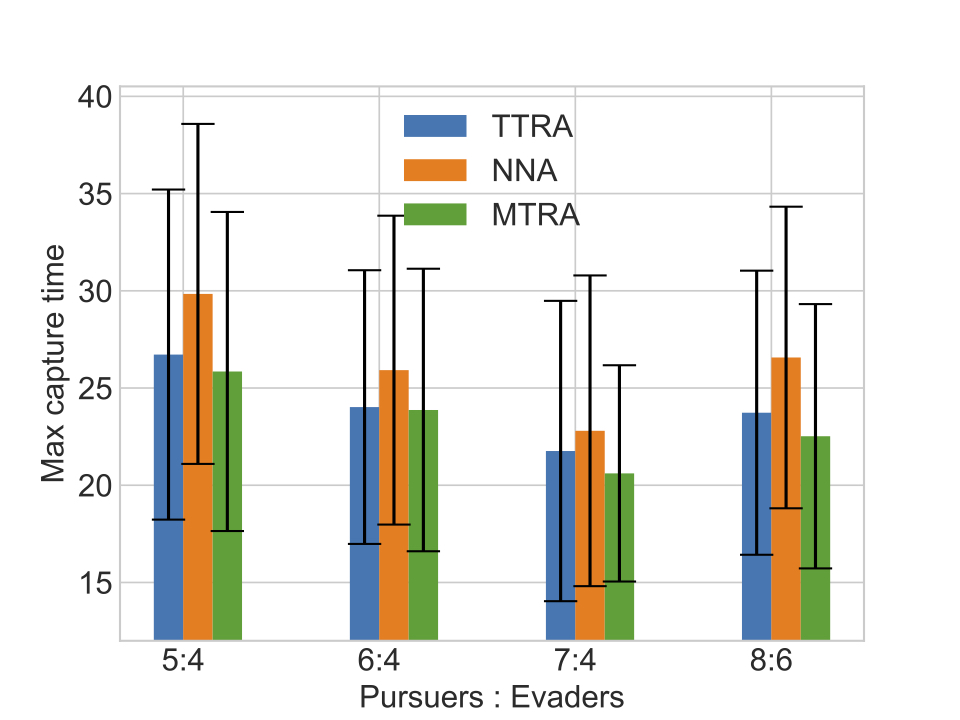}} \label{fig:comparison-max-AR0701SR}} 
	\centering
	\subfloat[Total capture time in `AR0414SR']
	{\includegraphics[width=0.245\textwidth, trim=10 5 46 40, clip=true]{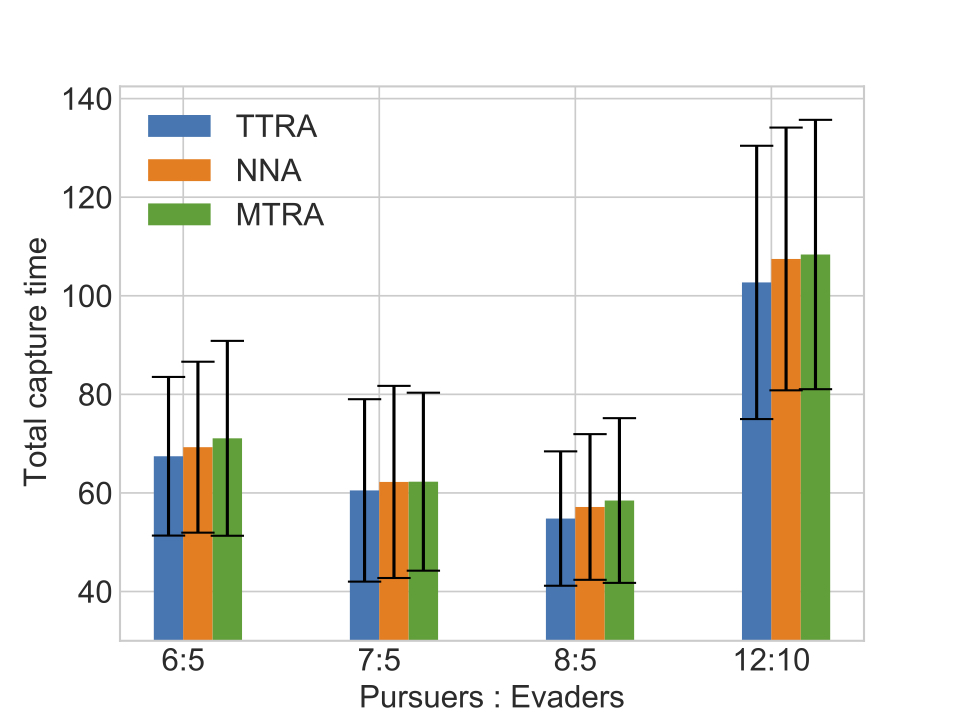} \label{fig:comparison-total-AR0414SR}} 
	\subfloat[Total capture time in  `AR0701SR']
	{{\includegraphics[width=0.245\textwidth, trim=10 5 46 40, clip=true]{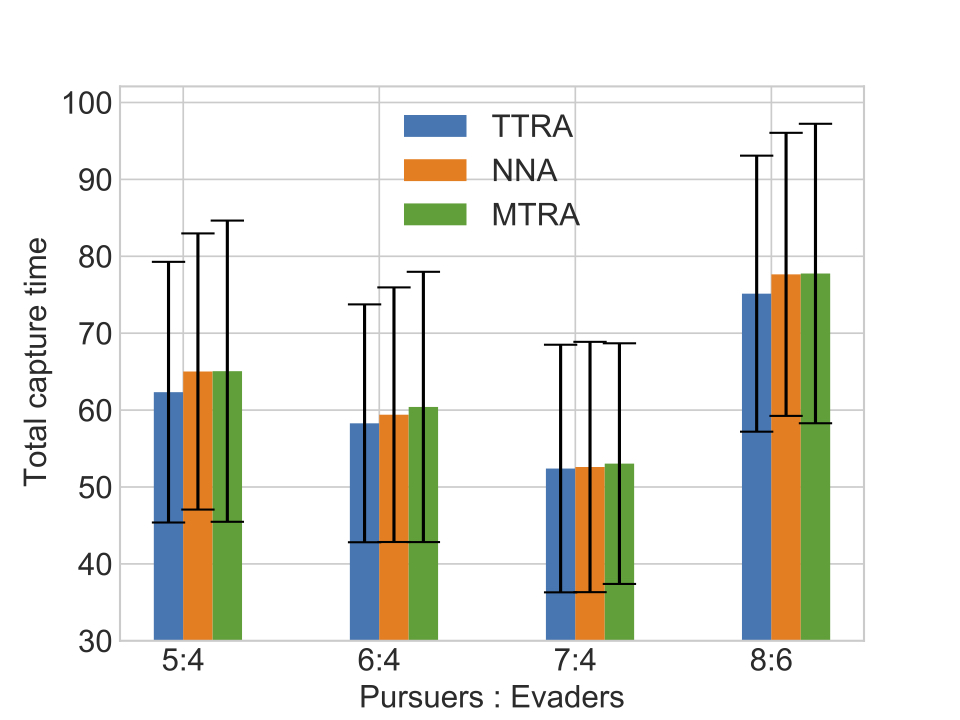}} \label{fig:comparison-total-AR0701SR}} 
	%
	\caption{Comparison of the average values of \changeJ{maximum} capture times (a-b) \changeJ{and total capture times (c-d) along with the} standard deviation in different environments and with different pursuer-to-evader ratios using the \changeJ{TTRA, NNA} and MTRA algorithms. Each bar represents data from $100$ simulations with randomized initial conditions.
	} 
	\label{fig:redundant_assignment_statistics}
\end{figure*}

Figure \ref{fig:redundant_assignment_statistics} shows a comparison between the proposed \changed{robot} assignment algorithms (TTRA and MTRA) and the NNA algorithm
for the \changed{aforementioned} environments.
From the comparison \changed{it is clear} that 
\changeJ{the MTRA algorithm consistently outperforms the other algorithms with respect to the maximum capture time (Figures~\ref{fig:comparison-max-AR0414SR}), while TTRA consistently outperforms the other algorithms with respect to the total capture time (Figures~\ref{fig:comparison-total-AR0414SR}).}
%
%
\changeC{In addition, Table \ref{tab:win_rate} shows win rates of TTRA and MTRA over NNA (for TTRA this is the proportion of simulations in which the total capture time for TTRA was lower than NNA, while for MTRA this is the proportion of simulations in which the total capture time for MTRA was lower than NNA). TTRA has a win rate of around 60\%, and MTRA has a win rate of over 70\%.}

\begin{table}[h] 
	\centering
	\begin{tabular}{c | c c c}
		\hline
		\changeJ{Algorithm Name} & \changeJ{AR0414SR} & \changeJ{AR0701SR} \\ [0.5ex]
		\hline
		\changeJ{TTRA} & \changeJ{69.3\%} & \changeJ{58.2\%}  \\
		\changeJ{MTRA} & \changeJ{78.0\%} & \changeJ{71.2\%} \\
		\hline
		
	\end{tabular}
\caption{\small \changeJ{Win rates of TTRA and MTRA algorithms over NNA. For a given set of initial conditions (initial position of pursuers and evaders), if TTRA takes less total time to capture all the evaders than NNA, it is considered a win for TTRA. While if MTRA takes less time to capture the last evader (maximum capture time) than NNA, it is considered as a win for MTRA.
}}
\label{tab:win_rate}
\end{table}

\changeJ{Clearly the advantage of the proposed greedy supermodular strategy for redundant robot assignment is \changeJ{statistically} significant.
\changed{Unsurprisingly, we also observe that}
increasing the number of pursuers tends to decrease the capture time.}

\changeJ{Figure \ref{fig:varying_noise} shows a comparison of the total and maximum capture times with varying measurement noise level (varying $k_{2}$) in the environment `AR0414SR' with a fixed number of pursuers and evaders, and with $20$ randomly generated initial conditions. 
As expected, higher noise leads to more capture time for all the algorithms. However MTRA still outperforms the other algorithms w.r.t. maximum capture time, while TTRA outperforms the other algorithms w.r.t. the total capture time.}
